\documentclass[runningheads,a4paper]{llncs}

\usepackage{times}
\usepackage{amssymb}
\usepackage{graphicx}
\usepackage{url}

\usepackage{siunitx}
\usepackage{hyperref}

\usepackage{booktabs} 
\usepackage{amsmath, amsfonts} 
\usepackage{subcaption} 
\usepackage{arydshln} 
\usepackage{tablefootnote} 
\usepackage{multirow} 
\usepackage[table,xcdraw]{xcolor} 
\usepackage{comment}
\usepackage{orcidlink} 

\newcommand{\keywords}[1]{\par\addvspace\baselineskip
\noindent\keywordname\enspace\ignorespaces#1}

\urldef{\mailsa}\path|{raul.monteiro, diogo.pernes}@priberam.pt|

\begin{document}

\title{Towards End-to-end Speech-to-text Summarization}

\titlerunning{Towards End-to-end Speech-to-text Summarization}

\author{Raul Monteiro\textsuperscript{1,2}\orcidlink{0009-0005-2435-1416} \and Diogo Pernes\textsuperscript{1,3}\orcidlink{0000-0001-5246-0402}}


\authorrunning{Raul Monteiro and Diogo Pernes}

\institute{Priberam, Alameda D.\ Afonso Henriques 41,
1000-123 Lisboa \\
\and
Instituto Superior Técnico, Universidade de Lisboa, Av.\ Rovisco Pais 1, 1049-001 Lisboa
\and
Universidade do Porto, Faculdade de Engenharia, R.\ Dr.\ Roberto Frias s/n, 4200-465 Porto \\
\mailsa
}

\index{Monteiro, Raul}
\index{Pernes, Diogo}

\toctitle{} \tocauthor{}

\maketitle

%
%
%
%
\begin{abstract}
Speech-to-text (S2T) summarization is a time-saving technique for filtering and keeping up with the broadcast news uploaded online on a daily basis. The rise of large language models from deep learning with impressive text generation capabilities has placed the research focus on summarization systems that produce paraphrased compact versions of the document content, also known as abstractive summaries. End-to-end (E2E) modelling of S2T abstractive summarization is a promising approach that offers the possibility of generating rich latent representations that leverage non-verbal and acoustic information, as opposed to the use of only linguistic information from automatically generated transcripts in cascade systems. However, the few literature on E2E modelling of this task fails on exploring different domains, namely broadcast news, which is challenging domain where large and diversified volumes of data are presented to the user every day. We model S2T summarization both with a cascade and an E2E system for a corpus of broadcast news in French. Our novel E2E model leverages external data by resorting to transfer learning from a pre-trained T2T summarizer. Experiments show that both our cascade and E2E abstractive summarizers are stronger than an extractive baseline. However, the performance of the E2E model still lies behind the cascade one, which is object of an extensive analysis that includes future directions to close that gap.
\keywords{Abstractive summarization, Speech-to-text summarization, End-to-end}
\end{abstract}

\section{Introduction}
\label{sec:introduction}

Broadcast news is mainly presented in large volumes of audio-visual multimedia, making it time-consuming to locate relevant information. S2T summarization systems help by identifying the most relevant content within human speech and producing a condensed form text suitable for the need. Extractive summarization selects relevant sentences or paragraphs from transcripts, but this method may sometimes lack cohesion and readability \cite{GUPTA201949}. The rise of large language models from deep learning has enabled reaching high-level understanding of input documents, besides having impressive text generation capabilities. Thus, the research focus has been recently placed on abstractive summarization systems, where the generated summaries are paraphrased compact versions of the speech content. They are more natural, coherent, and fluent, i.e. ideally similar to a summary written by a human specialist.\par
\begin{figure}[ht]
	\centering	\includegraphics[width=\linewidth]{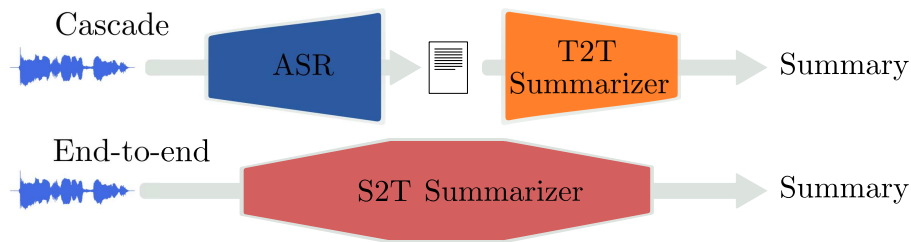}
	\caption{Illustration of a cascade and E2E system for S2T summarization.}
	\label{fig:cascade_vs_e2e}
\end{figure}\par
S2T summarization is usually achieved using a cascade approach (see Fig. \ref{fig:cascade_vs_e2e}), where an automatic speech recognition (ASR) model generates transcripts, followed by a text-to-text (T2T) summarization model that produces summaries \cite{DBLP:journals/corr/abs-2008-11897}. Deep learning, including attention-based architectures and self-supervised pre-training, has improved the performance of both models. Cascade abstractive systems using these components achieve strong results when trained on unpaired data for dialogue summarization tasks \cite{zhang-etal-2021-exploratory-study}. However, the transcripts produced by the ASR model may contain errors, so methods using confusion networks or language models have been proposed to improve robustness to these errors\cite{8683543,DBLP:journals/corr/abs-2006-01189}.\par 
Cascade systems used for S2T summarization fail to utilize non-verbal and acoustic information that could be useful for summarization \cite{tundik19_interspeech}. E2E modelling (see Fig. \ref{fig:cascade_vs_e2e}) has been proposed to address this issue in two different articles \cite{DBLP:journals/corr/abs-2110-06263,matsuura2023leveraging}. These systems do not make use of an intermediate speech recognition step and instead jointly optimise an acoustic and language model. However, E2E modelling requires large amounts of paired audio/summary data and the scarcity of publicly available large corpora on the broadcast news domain requires techniques to leverage external data.\par 
This work proposes both a cascade and novel E2E models for S2T abstractive summarization of broadcast news. The former uses fine-tuned ASR and T2T abstractive summarizer on a broadcast news dataset. The E2E system follows the encoder-decoder paradigm and utilizes speech features extracted using a self-supervised pre-trained speech representation model as input \cite{evain2021task}. It leverages external data from text corpora through transfer learning from a T2T abstractive summarizer. Both models are compared against an extractive cascade baseline and to each other using ROUGE scores and human evaluation. We release our source code publicly\footnote{\url{https://github.com/Priberam/S2TSumm}}.\par
The remainder of this paper is organized as follows: in section \ref{sec:related}, we present the related work; in section \ref{sec:dataset}, we propose a new corpus of broadcast news in French, which is used to evaluate the models developed in this work; in section \ref{sec:models}, we describe the architectures of the cascade and novel E2E abstractive summarizers, how the latter benefits from the former through transfer learning and we introduce an extractive baseline; in section \ref{sec:adapter}, we detail the architecture and pre-training of the cross-modal adapter, which is the encoder of the E2E S2T abstractive summarizer that maps speech to textual features; in section \ref{sec:evaluation}, we present the results for automatic and human evaluations; in section \ref{sec:discussion}, we discuss the obtained results; section \ref{sec:conclusion} concludes this work  and includes future directions for improving the performance of the E2E model.
\section{Related Work}
\label{sec:related}

\paragraph{Automatic Speech Representation Learning and Recognition:}

Wav2vec 2.0 (W2V2) \cite{DBLP:journals/corr/abs-2006-11477} is a transformer-based encoder-only model for extracting deep representations from raw audio waveforms, which was trained with self-supervised objectives. Evain et al.\ \cite{evain2021task} found that speech representations extracted from W2V2 models trained on French data lead to better performance on several speech-related tasks than using human-tailored speech features like Mel filter bank (MFB) and Mel-frequency cepstral coefficients features (MFCC) \cite{Davis1980ComparisonOP,Furui1986SpeakerindependentIW}. Pasad et al.\ \cite{DBLP:journals/corr/abs-2107-04734} used a metric called Projection Weighted Canonical Correlation Analysis (PWCCA) \cite{https://doi.org/10.48550/arxiv.1806.05759} to study the layer representations of W2V2 models. PWCCA was applied to compare the learnt representations of each layer with external representations, for instance, MFB features and GloVe word embeddings \cite{pennington2014glove}. It is uncovered that the pre-trained W2V2 models encode more semantic information in inner layers, whereas acoustic information is mostly represented in the outer layers.\par
W2V2 and its variants can be used as pre-initializations and directly trained for speech recognition using the Connectionist Temporal Classification (CTC) objective. The literature also contains fully supervised approaches like Whisper \cite{radford2022whisper}, which was jointly trained for ASR and speech translation using very large amounts of data crawled from the web.

\paragraph{Text-to-text Abstractive Summarization:}

Most state-of-the-art approaches for T2T abstractive summarization make use of pre-trained large sequence-to-sequence (Seq2seq) language models like BART and fine-tune them on abstractive summarization datasets \cite{DBLP:journals/corr/abs-1910-13461}. Rothe et al.\ \cite{DBLP:journals/corr/abs-1907-12461} proposed an alternative approach in which encoder-only language models like RoBERTa \cite{DBLP:journals/corr/abs-1907-11692} could be promoted to decoder modules, and the resulting encoder-decoder models could be fine-tuned for downstream tasks.\par

\paragraph{End-to-end Speech-to-text Abstractive Summarization:}

To the best of our knowledge, only two works directly exploit E2E modelling of S2T summarization. Sharma et al.\ \cite{DBLP:journals/corr/abs-2110-06263} used a restricted self-attention to enable processing long input audios with a transformer architecture. The authors first trained a randomly initialized model for ASR, and then trained it for S2T abstractive summarization using a 2000h corpus of instructional videos. Matsuura et al.\ \cite{matsuura2023leveraging} further leveraged a T2T abstractive summarization corpus using a text-to-speech voice synthesizer as a way of data augmentation. Both papers report better results than strong cascade baselines.\par
\section{Dataset}
\label{sec:dataset}

For this work, we built a dataset for S2T abstractive summarization of broadcast news in French, that was built from articles that can be found in the EuroNews website\footnote{\url{https://www.euronews.com/about}}. Each news article from EuroNews has an audio, an abstractive summary of the news content and the article body. Since the latter is not always a perfect transcript of the audio, we employed an automatic procedure for selecting the news articles whose article bodies are perfect (or almost perfect) transcripts of the audios. An XLSR-based ASR model\footnote{\url{https://huggingface.co/facebook/wav2vec2-large-xlsr-53-french}} was used to produce artificial transcripts from the audios. Afterwards, the word error rate (WER) evaluation metric was applied between the automatically generated transcript and the article body. A threshold for the WER of \num{45}{\%} was set, such that articles associated with higher values of WER were discarded. The remaining articles were randomly shuffled and separated into three distinct splits with sizes of \num{13380}, \num{1672} and \num{1673} for the \textit{train}, \textit{dev} and \textit{test} splits, respectively, and this final corpus was named BNews\footnote{We are in contact with EuroNews to have a public license of this dataset.}. The mean audio duration per article is about \SI{87}{s}.
\section{Model Architectures}
\label{sec:models}

\begin{figure}[ht]
	\centering	\includegraphics[width=\linewidth]{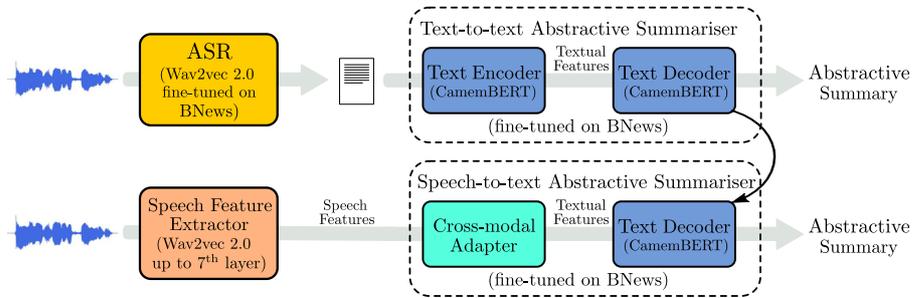}
	\caption{Architectures of the cascade and E2E abstractive summarizers.}
	\label{fig:architectures}
\end{figure} 

\subsection{Cascade}

The cascade abstractive summarizer requires both an ASR system and a T2T abstractive summarizer. Fig. \ref{fig:architectures} illustrates the realisation of the cascade and E2E abstractive summarizers.\par

\paragraph{Automatic Speech Recognizer}

The ASR model was built from a W2V2 model\footnote{\url{https://huggingface.co/LeBenchmark/wav2vec2-FR-7K-base}} that was pre-trained on French speech data. The pre-trained model was loaded to a \texttt{Wav2Vec2ForCTC} object from the \texttt{Transformers} library of Huggingface\footnote{\url{https://huggingface.co/docs/transformers/index}}. This model consists of a pre-trained W2V2 model, followed by a linear layer and a \texttt{softmax}. The model is trained for speech recognition using the French sub-dataset of the Common Voice Corpus 10.0 (CV) with the CTC objective. The vocabulary contains \num{222} characters extracted from the \textit{dev} split of the BNews corpus. The model was further fine-tuned on the latter from the checkpoint that showed lower WER on the \textit{dev} split of the CV. The WER on the \textit{test} split of the BNews corpus was (\num[separate-uncertainty]{18.8(3)})\:\%, where the \texttt{BasicTextNormalizer} from Whisper\footnote{\url{https://github.com/kurianbenoy/whisper_normalizer}} was used for text normalization.

\paragraph{Text-to-text Abstractive Summarizer}

A publicly available pre-trained T2T abstractive summarizer\footnote{\url{https://huggingface.co/mrm8488/camembert2camembert_shared-finetuned-french-summarization}} is used as the starting initialisation for the model weights. The author of this model built it from two CamemBERTs \cite{DBLP:journals/corr/abs-1911-03894}, following the technique introduced by Rothe et al.\ \cite{DBLP:journals/corr/abs-1907-12461}, and trained it for abstractive summarization using the French sub-dataset from the MLSUM corpus \cite{scialom-etal-2020-mlsum}. The summarizer was further fine-tuned on the BNews training data. Only the weights of the decoder are updated during this fine-funing. The checkpoint that showed maximum \texttt{ROUGE-2} score on the \textit{dev} split of the BNews corpus was selected.

\subsection{End-to-end}

The novel E2E implementation for S2T abstractive summarization proposed in this work does not directly use the audio waveform or MFB/MFCC features as input. Instead, it takes speech features generated by the same pre-trained W2V2 model that was trained for ASR. The S2T abstractive summarizer takes the speech features and converts them to a summary of the audio content.

\paragraph{Speech Feature Extractor}

Following the same methodology used in \cite{DBLP:journals/corr/abs-2107-04734}, we computed the PWCCA scores\footnote{\url{https://github.com/google/svcca}} between word-level embeddings extracted from each transformer layer of the W2V2 base model and pre-trained French word embeddings, which were obtained in \cite{ferreira-etal-2016-jointly}. It is found that the 7\textsuperscript{th} transformer layer is the one that generates representations more similar to word embeddings. For this reason, the speech feature extractor is composed of all the layers of the W2V2 model up to and including the 7\textsuperscript{th} transformer layer.

\paragraph{Speech-to-text Abstractive Summarizer}

As is illustrated on Fig. \ref{fig:architectures}, the decoder from the T2T summarizer is transferred to the S2T summarizer, which allows leveraging the MLSUM training data. The sequences of speech features do not lie in the same representation space as the textual features that the encoder of the T2T summarization model generates. For that reason, one must add an additional module that bridges the speech feature extractor and the decoder. This encoder is responsible for mapping sequences of audio features to sequences of textual features, and shall be hereby denoted as cross-modal adapter. The particular architecture of the cross-modal adapter and its pre-training are the subjects of section \ref{sec:adapter}. After being pre-trained, the whole S2T abstractive summarizer is fine-tuned on the BNews corpus, and the checkpoint with higher \texttt{ROUGE-2} score on the \textit{dev} split is selected for evaluation.

\subsection{Extractive Baseline}

The extractive baseline uses the same ASR system as the cascade abstractive summarizer. We adopted a simple centroid-based approach, where the sentence embeddings were provided by a publicly available unsupervised extractive CamemBERT-based model\footnote{\url{https://github.com/ialifinaritra/Text_Summarization}}. The summary is constructed by concatenating the top-$k$ closest sentences to the centroid until a maximum number of words $\Bar{w}$ is reached, where $\Bar{w}=24$ was set to match the average length of the \textit{dev} split of the BNews corpus.
\section{Cross-modal Adapter}
\label{sec:adapter}

Given a dataset $\mathcal{D}=\{(w^{(i)},t^{(i)},r^{(i)} )\}_{i=1}^N$ made of triplets of an audio waveform $w^{(i)}$, the speech transcript $t^{(i)}$ and reference summary $r^{(i)}$, the speech feature extractor generates an 
$L^{(i)}$-sized sequence of speech features $x^{(i)}=\{x^{(i)}_j\}_{j=1}^{L^{(i)}}$, $x^{(i)}_j\in\mathbb{R}^{768}$, from the audio waveform $w^{(i)}$, whereas the encoder from the T2T summarizer extracts a sequence of textual features or embeddings $y^{(i)}=\{y^{(i)}_j\}_{j=1}^{T^{(i)}}$, $y^{(i)}_j\in\mathbb{R}^{768}$ from the speech transcript $t^{(i)}$. The cross-modal adapter must be developed for mapping from sequences $x^{(i)}$ to $y^{(i)}$ (see Fig.\ \ref{fig:architectures}).

\subsection{Architecture}

The architecture of the cross-modal is encoder-decoder. The encoder is a 1-layer BiLSTM and the decoder is a 1-layer forward LSTM, both of hidden dimension \num{768}. Since the speech feature extractor generates speech features with an output frequency of \SI{50}{\hertz}, whereas spoken text roughly contains 2-3 words per second, the encoder BiLSTM is preceded by a 2-layer convolutional neural network to reduce the length of the sequence of speech features from $L$ to $\Tilde{L}\approx L/4$. The cross-modal adapter contains several attention mechanisms, which follow closely the work done in \cite{DBLP:journals/corr/PaulusXS17}. Table \ref{tab:cross_modal_adapter_attentions} summarizes those attention mechanisms. The $\overrightarrow{h}_i^\mathrm{e}$, $\overleftarrow{h}_i^\mathrm{e}$ and $h_t^\mathrm{d}$ stand for encoder forward, encoder backward and decoder LSTM hidden states, respectively. The $t$-th textual embedding $y_t$ is just a linear projection $y_t=W_{text}[h^\mathrm{d}_t\lVert s^\mathrm{d}_t\lVert c_t^\mathrm{e}\lVert c_t^{d}]$, where $s_t^\mathrm{d}$ is the decoder LSTM cell state and $[\cdot\lVert\cdot]$ denotes vector concatenation.\par 
\begin{table}[ht]
	\centering
        \footnotesize
	\begin{tabular}{c c c c}
		\toprule
		\textbf{Attention} & 
		\textbf{Intra-temporal Cross}& 
		\textbf{Intra-decoder}& 
		\textbf{EOS Generation}
		\\
		\midrule
		\textbf{Query} 
		& $h_t^\mathrm{d}$
            & $h_t^\mathrm{d}$
            & $y_t$ \\ \midrule
            \textbf{Keys}
            & $\{h_i^\mathrm{e}=[\overrightarrow{h}_i^\mathrm{e}\lVert \overleftarrow{h}_i^\mathrm{e}]\}_{1\leq i\leq \Tilde{L}}$
            & $\{h^\mathrm{d}_{t'}\}_{t'<t}$
            & $\{y_{t'}\}_{t-w\leq t'\leq t+w}$ \\ \midrule
            \textbf{Values}
            & $\{h_i^\mathrm{e}=[\overrightarrow{h}_i^\mathrm{e}\lVert \overleftarrow{h}_i^\mathrm{e}]\}_{1\leq i\leq \Tilde{L}}$
            & $\{h^\mathrm{d}_{t'}\}_{t'<t}$
            & $\{y_{t'}\}_{t-w\leq t'\leq t+w}$ \\ \midrule
            \multirow{2}{*}{\textbf{En. Scores}}
            & $e_{ti}^\mathrm{e}={h_t^\mathrm{d}}^TW_{\mathrm{attn}}^\mathrm{e}h_i^\mathrm{e}$
            & \multirow{2}{*}{$e_{tt'}^\mathrm{d}={h_t^\mathrm{d}}^TW_{\mathrm{attn}}^\mathrm{d}h_{t'}^\mathrm{d}$}
            & \multirow{2}{*}{$e_{tt'}^{\mathrm{eos}}={y_{t'}}^TW_{\mathrm{attn}}^{\mathrm{eos}}y_t$} \\ 
            & ${e'}^\mathrm{e}_{ti}=\frac{\exp(e^\mathrm{e}_{ti})}{\sum_{j=1}^{t-1}\exp(e^\mathrm{e}_{ji})}$ & \\
            \midrule
            \textbf{Att. Weights}
            & $\alpha^\mathrm{e}_{ti}=\frac{{e'}^\mathrm{e}_{ti}}{\sum_{j=1}^{L}{e'}^\mathrm{e}_{tj}}$
            & $\alpha^\mathrm{d}_{tt'}=\frac{\exp(e^\mathrm{d}_{tt'})}{\sum_{j=1}^{t-1}\exp(e^\mathrm{d}_{tj})}$
            & $\alpha^{\mathrm{eos}}_{tt'}=\frac{\exp(e^{\mathrm{eos}}_{tt'})}{\sum_{j=t-w}^{t+w}\exp(e^{\mathrm{eos}}_{tj})}$ \\ \midrule 
            \textbf{Cont. Vectors}
		& $c_t^\mathrm{e}=\sum_{i=1}^{\Tilde{L}}\alpha_{ti}^\mathrm{e}h_i^\mathrm{e}$
            & $c_t^\mathrm{d}=\sum_{t'=1}^{t-1}\alpha_{tt'}^\mathrm{d}h_{t'}^\mathrm{d}$
            & $c_t^\mathrm{eos}=\sum_{t'=t-w}^{t+w}\alpha_{tt'}^\mathrm{eos}y_{t'}$ \\
		\bottomrule
	\end{tabular}
	\caption{Different attention mechanisms used in the cross-modal adapter.}	\label{tab:cross_modal_adapter_attentions}
\end{table}\par 
Textual embeddings are continuously-valued on high-dimensional spaces. As such, there is not a direct way to stop the generation process at inference time. This problem is circumvented by training an additional attention mechanism and a neural layer for predicting the end of sequence. Given a set $\{\Hat{y}_t\}_{t=1}^{T}$ of $T$ (fixed) predicted textual embeddings, a restricted attention mechanism is applied on every $\Hat{y}$ using a window of size $w$, which was set to \num{1} in all experiments. Details of the attention mechanism can be found in table \ref{tab:cross_modal_adapter_attentions}. A linear layer followed by a \texttt{sigmoid} function $\sigma(\cdot)$ are used to obtain the probability $p^{\mathrm{eos}}_t=\sigma\left(W_{\mathrm{eos}}[h^\mathrm{d}_t\lVert s^\mathrm{d}_t\lVert c_t^{\mathrm{eos}}]\right)\in [0\text{,}\:1]$ of reaching the end of the sequence at time step $t$.\par
At inference time, the decoder of the cross-modal adapter auto-regressively generates $T=512$ textual embeddings $\Hat{y}=\{\Hat{y}_t\}_{t=1}^{T}$. For each textual embedding, a corresponding probability $\Hat{p}^{\mathrm{eos}}_t$ of having reached the end of sequence is associated. A straightforward method to choose the end of sequence is to find the first instant $t_{\pi}$ such that $\Hat{p}^{\mathrm{eos}}_{t_{\pi}}>\pi$, where $\pi\in[0,1]$ is a probability threshold. It was set to \num{0.5} in all experiments. Finally, the cross-modal adapter outputs a reduced sequence of textual embeddings $\Hat{y}^{\mathrm{red}}=\{\Hat{y}_t\}_{t=1}^{t_{\pi}}$.\par 

\subsection{Pre-training}
\label{sec:pretraining}

The pre-training of the cross-modal adapter encompasses three controlled steps, which are described below. The input speech features and target textual features were normalized such that each dimension had zero mean and unit variance.\par
Stage 1: At this stage, we used the same Common Voice corpus that was used to train the ASR model. A proportion of speech features from the sequence $x^{(i)}$ is randomly masked, where for every element of the sequence there is a probability $p_{\mathrm{mask}}=\num{6.5e-2}$ of starting a masked span at that position with length $M_{\mathrm{mask}}=\num{10}$ (values identical to the ones used to train the W2V2 model). The cross-modal adapter is trained to minimise the mean squared error (MSE) between the reference embeddings $y^{(i)}$ and the ones predicted from the masked sequence $\Hat{y}^{(i)}$.\par
Stage 2: We dropped the CV dataset and used the BNews corpus during this training stage. The objective remains to minimizing the MSE. Masking is no longer used and the default teacher forcing algorithm for training Seq2seq models is replaced by the peeling back algorithm introduced in \cite{alternatives_to_teacher_forcing}. For the $j$-th minibatch or training step, we use linear decay for the teacher forcing ratio $\lambda(j)=\max(\epsilon,k-cj)$, where $\epsilon=\num{5.0e-1}$, $k=1.0$, and $c=\num{8.0e-6}$.\par
Stage 3: The cross-modal adapter is now trained to predict the end of the sequence of textual embeddings, again using the BNews dataset. Given a $T^{(i)}$-sized sequence of predicted textual embeddings $\Hat{y}^{(i)}$, predicting for every $\Hat{y}^{(i)}_t$ whether it is the end of the sequence is a binary classification problem. Minimizing a binary cross-entropy loss suffices. All the model weights are frozen except for the ones directly associated with end-of-sequence prediction ($W^{\mathrm{eos}}_{\mathrm{attn}}$ and $W_{\mathrm{eos}}$), and one also makes use of the peeling back algorithm with linear decay, where $\epsilon=0.0$, $k=1.0$ and $c=\num{3.0e-4}$.\par
After this three-stage pre-training, the cross-modal adapter and the text decoder are jointly trained for abstractive summarization using the BNews dataset in a multitask objective consisting of the usual cross-entropy loss for summarization and the binary cross-entropy for EOS detection.
\section{Evaluation}
\label{sec:evaluation}

\subsection{Automatic Evaluation}

For assessing the performance of the different implementations developed in this work, we make use of the ROUGE package\footnote{https://huggingface.co/spaces/evaluate-metric/rouge}, more specifically, the \texttt{ROUGE-1}, \texttt{ROUGE-2}, \texttt{ROUGE-L} and \texttt{ROUGE-Lsum} metrics. The decoding for the cascade and E2E abstractive summarizers is performed with beam search. Table \ref{tab:final_comparison} compares the ROUGE scores for the extractive baseline and both cascade and E2E abstractive summarizers on the \textit{test} split of the BNews corpus. We include the topline performance, which is simply the T2T abstractive summarizer from the cascade system applied on the gold transcripts (GT), and thus serves as an upper bound for the performance of the cascade abstractive summarizer. We also performed ablation studies for the following cases: the S2T abstractive summarizer is not fine-tuned on the BNews corpus after the pre-training of the cross-modal adapter (nFT); there is no fine-tuning and the cross-modal adapter additionally does not make use of its predictions for the end-of-sequence positions of the sequences of textual embeddings and uses instead the gold ones (G-EOS); the pre-training of cross-modal adapter described in section \ref{sec:pretraining} is not performed and the S2T abstractive summarizer is directly trained using the BNews dataset (nPre).\par 
\begin{table}[ht!]
	\centering
        \setlength{\tabcolsep}{4pt}
        \footnotesize
        \begin{tabular}{l|cccc}
		\textbf{Model} & \texttt{ROUGE-1} & \texttt{ROUGE-2} & \texttt{ROUGE-L} & \texttt{ROUGE-Lsum} \\
            \hline
		Topline (GT + T2T) & \num[separate-uncertainty]{45.9(14)} & \num[separate-uncertainty]{33.0(18)} & \num[separate-uncertainty]{39.7(16)} & \num[separate-uncertainty]{41.6(14)} \\ \hdashline
            Cascade (ASR + T2T) & \num[separate-uncertainty]{41.6(12)} & \num[separate-uncertainty]{26.2(14)} & \num[separate-uncertainty]{35.7(12)} & \num[separate-uncertainty]{37.6(12)} \\
            \hdashline
            E2E & \num[separate-uncertainty]{37.8(12)} & \num[separate-uncertainty]{23.7(12)} & \num[separate-uncertainty]{32.8(12)} & \num[separate-uncertainty]{33.9(12)} \\
            E2E (nFT) & \num[separate-uncertainty]{30.0(10)} & \num[separate-uncertainty]{16.1(10)} & \num[separate-uncertainty]{25.9(10)} & \num[separate-uncertainty]{26.6(10)} \\
            E2E (G-EOS) & \num[separate-uncertainty]{29.8(10)} & \num[separate-uncertainty]{15.9(10)} & \num[separate-uncertainty]{25.7(10)} & \num[separate-uncertainty]{26.5(10)} \\
            E2E (nPre) & \num[separate-uncertainty]{16.8(4)} & \num[separate-uncertainty]{2.4(2)} & \num[separate-uncertainty]{12.6(3)} & \num[separate-uncertainty]{13.2(3)} \\
            \hdashline
            Extractive & \num[separate-uncertainty]{23.8(8)} & \num[separate-uncertainty]{8.3(8)} & \num[separate-uncertainty]{17.9(8)} & \num[separate-uncertainty]{18.8(8)} \\
	\end{tabular}
	\caption{Comparison between the ROUGE scores for the topline, baseline, cascade and end-to-end (E2E) on the \textit{test} split of the BNews corpus. We show results without the final fine-tuning of the S2T abstractive summarizer (nFT), when using the ground truth end-of-sequence positions (G-EOS) and without the pre-training (nPre) of the cross-modal adapter. Every score is provided with a \num{95}{\%} confidence interval for the mean.}
 \label{tab:final_comparison}
\end{table} 
All the abstractive systems outperform the extractive baseline, which was expected given that the target summaries from our corpus are abstractive. The cascade abstractive summarizer yields worse scores than the topline model, which is due to ASR error propagation. On the other hand, the E2E model performs worse than the cascade model, as measured by ROUGE scores. This contrasts with the fact that, theoretically, E2E modelling allows leveraging non-verbal and acoustic information besides the linguistic one from transcripts, which is the only type of information that cascade systems have access to. Regarding the ablation studies, by comparing the performance of the E2E and E2E (nFT) models, it is found that fine-tuning the S2T abstractive summarizer after the pre-training of the cross-modal adapter significantly improves the ROUGE scores with a relative increase on the interval of \num{25}{\%} - \num{50}{\%}. The similarity between the ROUGE scores of the E2E (nFT) and E2E (G-EOS) models allows us to conclude that the cross-modal adapter performs equally well either when using its own predictions for the end-of-sequence positions of the sequences of textual embeddings or when using the ground truth ones. Finally, the gap between E2E and E2E (nPre) proves that the proposed pre-training of the cross-modal adapter provides a very significant performance increase.

\subsection{Human Evaluation}

ROUGE metrics are simple automatic methods to evaluate the overlap between predicted and reference summaries. However, these metrics alone fail to evaluate important features like factual consistency (FC), relevance (R) and fluency (F). To evaluate these attributes, following the same procedure and criteria definition as in \cite{pernes-etal-2022-improving}, we do pairwise comparisons between the summaries generated by the extractive baseline, the cascade and E2E systems. Given an entry of the dataset, we (one of the authors) were provided with the gold transcript and every pairwise combination of the summaries generated by the three systems. Afterwards, we were asked to rank the generated summaries according to the three criteria. For each criterion, we would evaluate whether the first summary is better than the second, tied with, or worse than the second summary. To make the evaluation process as unbiased as possible, the names of the models that generated each summary were not shown and the order with which they appeared was randomized. We randomly selected \num{30} examples from the \textit{test} split of the BNews corpus and table \ref{tab:human_evaluation} shows the proportion of times that each system was considered the best for every pairwise comparison, according to each criterion. We show two examples of the evaluated summaries on table \ref{tab:examples}.\par 
\begin{table}[ht!]
	\centering
	\renewcommand{\arraystretch}{1.1}
        \footnotesize
        \begin{tabular}{l|ccc}
		 & \textbf{FC} & \textbf{R} & \textbf{F} \\
        \hline
		Extractive is better & \num{0.17} & \num{0.10} & \num{0.13} \\
		Tie & \textbf{0.73} & \num{0.07} & \textbf{0.47} \\
		Cascade (ASR + T2T) is better & \num{0.10} & \textbf{0.83} & \num{0.40} \\ 
		\hline
		Extractive is better & \textbf{0.63} & \num{0.23} & \num{0.30} \\
		Tie & \num{0.30} & \num{0.13} & \num{0.33} \\
		End-to-end is better & \num{0.07} & \textbf{0.63} & \textbf{0.37} \\
		\hline
		Cascade (ASR + T2T) is better & \textbf{0.60} & \textbf{0.57} & \num{0.37} \\
		Tie & \num{0.40} & \num{0.33} & \textbf{0.53} \\
		End-to-end is better & \num{0.00} & \num{0.10} & \num{0.10} \\
	\end{tabular}
	\caption{Proportion of times that each model was considered the best in each pairwise comparison, according to each criterion with respect to factual consistency (FC), relevance (R) and fluency (F).}
 \label{tab:human_evaluation}
\end{table}
The extractive baseline has been found to be very strong regarding factual consistency. This is consistent with the fact that extractive summaries are directly made of segments from automatically generated transcripts, and therefore factual inconsistencies may only come from ASR misspellings or unfortunate concatenation of sentences that together change the meaning of the original content. The cascade system is competitive against the extractive baseline in terms of factual consistency, but the E2E system performs very poorly on that attribute. Regarding relevance, the cascade and E2E systems are found to perform better than the extractive baseline. This was expected, since extractive summaries contain whole sentences that may include irrelevant information, or there may not exist sentences that give a comprehensive overview of the whole news. The cascade system also dominates the fluency attribute, and although the E2E model is generally more fluent than the extractive system, the difference is not as large as expected. The second example provided on table \ref{tab:examples} is illustrative of the cases when the E2E model generates a repetitive summary, therefore compromising its fluency. When comparing only the abstractive summarizers, the cascade and E2E ones, we clearly see that the cascade system produces summaries that are better in all the three evaluated attributes, which is in line with the automatic evaluation with the ROUGE metrics.\par 

\begin{table}[ht]
\begin{scriptsize}
\begin{tabular}{p{1cm}|p{11cm}}
       & Text \\ \hline
        Transcript & Des milliers de personnes rassemblées à Madrid pour dire "non" à la grâce des indépendantistes catalans, envisagée par le chef du gouvernement espagnol. En Espagne, des milliers de personnes se sont rassemblées ce dimanche à Madrid pour dire "non" à la grâce des indépendantistes catalans. (...) \\ 
        \hdashline[0.5pt/5pt]
        Reference  
        & Des milliers de personnes rassemblées à Madrid pour dire Non à la grâce des indépendantistes catalans, envisagée par le chef du gouvernement espagnol.  \\
        \hdashline[0.5pt/5pt]
        Extractive  
        & Ces deux partis sont profondément oposés à l'initiative de l'actuele Premier ministre, Le socialiste Pedron Sanchez, qui souhaite acorder la grâce au leader séparatistes condamné pour la tentative de sécesion de 2017  \\ \hdashline[0.5pt/5pt]
        Cascade    
        & En Espagne, des miliers de persones se sont rasemblées ce dimanche à Madrid pour dire non à la grâce des indépendantistes catalans.  \\ 
        \hdashline[0.5pt/5pt]
        E2E    
        & Des milliers de personnes se sont rassemblées ce dimanche à Madrid pour dire non à la grâce des indépendantistes catalans.  \\ 
        \hline       
        Transcript & La Belgique se dit prête à accueillir Jean-Pierre Bemba. Ce jeudi, le chef de la diplomatie belge, Didier Reynders, a annoncé que le pays était prêt à accueillir l'ancien dirigeant congolais, qui souhaite rejoindre sa famille, qui vit dans le pays. (...) \\ 
        \hdashline[0.5pt/5pt]
        Reference  
        & Le chef de la diplomatie belge l'a annoncé ce jeudi. \\
        \hdashline[0.5pt/5pt]
        Extractive  
        & Cet ancien home d'afaires devenue ensuite vice-président de la République démocratique du Congo, avait été condamné en première instance à 18 ans de prison il y a deux ans \\
        \hdashline[0.5pt/5pt]
        Cascade    
        & Didier Renders a finalement été acquité par la Cour pénale internationale, des acusations de crime de guere et de crimes contre l'humanité la semaine dernière qui a permis la mise en liberté conditionele de celui qui a été incarcéré pendant dix ans à La Haye. \\
        \hdashline[0.5pt/5pt]
        E2E
        & Le chef de la diplomatie belge a annoncé que le pays était prêt à accueillir Jean-Claude Juncker. Ce jeudi, le chef de la diplomatie belge a annoncé que le pays était prêt à accueillir Jean-Claude Juncker. \\
        \hline
\end{tabular}
\end{scriptsize}
\caption{Examples of summaries produced by the different summarizers.}
\label{tab:examples}
\end{table}
\section{Discussion}
\label{sec:discussion}

The results from automatic and human evaluation point out that the E2E abstractive summarizer underperforms with respect to the cascade one. This underperformance may be explained if one considers the several sub-modules of the cascade and E2E summarizers. Both make use of a W2V2-based model either for speech recognition or plain speech feature extraction. The T2T abstractive summarizer of the cascade system and the S2T abstractive summarizer of the E2E system share the same decoder, but differ strongly on the encoder. Thus, the limited performance of the proposed novel E2E implementation when compared with the cascade system must be sourced on the particular realization of the cross-modal adapter.
We have strong reasons to believe that the large T2T summarization corpus (MLSUM \cite{scialom-etal-2020-mlsum}), to which the encoder of the T2T summarizer was exposed during its training for abstractive summarization, played a significant role. It is likely that this enormous amount of external data makes the text encoder generate much richer textual latent representations than the ones the cross-modal adapter could possibly generate, given that it only had access to the summarization training data from the BNews corpus during its development.\par

\section{Conclusion}
\label{sec:conclusion}

We proposed a novel E2E model for S2T abstractive summarization of broadcast news in French. It leverages external data from T2T summarization corpora through transferring the decoder from a T2T abstractive summarizer. Additionally, we proposed a clever pre-training of the cross-modal adapter that leverages external data from an ASR dataset besides the BNews corpus. We presented an extensive analysis that took into account automatic and human evaluations for assessing the quality of the generated summaries. Although the E2E model did not beat the cascade, our contributions helped to close the performance gap between the two approaches, as is shown by our ablation studies.\par
The low amount of abstractive summarization training data for pre-training the cross-modal adapter has been shown as the most likely source of the underperformance of E2E model. Future work should focus on enriching the training of the cross-modal adapter. For instance, by also transferring the text encoder from the T2T abstractive summarizer and carefully train it to process speech features as input. Another possible and not mutually exclusive direction would be the use of augmented data from T2T summarization corpora through speech synthesis to enlarge the training data. Finally, the lack of large corpora with speech/summary pairs severely jeopardizes any fully supervised approach for developing an E2E system. Future work on developing this kind of datasets is needed in order to improve the promising E2E systems.\par

\subsubsection*{Acknowledgments} 

This work was supported by the EU H2020 SELMA project (grant agreement No. 957017).

\bibliographystyle{splncs04}
\bibliography{paper}

\end{document}